\documentclass[10pt,twocolumn,letterpaper]{article}

\usepackage{cvpr}
\usepackage{times}
\usepackage{epsfig}
\usepackage{graphicx}
\usepackage{amsmath}
\usepackage{amssymb}

\usepackage[pagebackref=true,breaklinks=true,colorlinks,bookmarks=false]{hyperref}

\cvprfinalcopy 

\def\cvprPaperID{****} 

\ifcvprfinal\fi
\begin{document}
	
	\title{Probabilistic Global Scale Estimation for MonoSLAM\\ Based on Generic Object Detection}
	
	\author{Edgar Sucar, Jean-Bernard Hayet\\
		Centro de Investigaci\'on en Matem\'aticas - Universidad de Guanajuato\\
		Jalisco S/N, Col. Valenciana CP: 36023 Guanajuato, Gto, México\\
		{\tt\small \{edgar.sucar@cimat.mx, jbhayet@cimat.mx\}}
	}
	
	\maketitle
	
	\begin{abstract}
This paper proposes a novel method to estimate the global scale of a 3D reconstructed model within a Kalman filtering-based monocular SLAM algorithm. Our Bayesian framework integrates height priors over the detected objects belonging to a set of broad predefined classes, 
based on recent advances in fast generic object detection. Each observation is produced on single frames, so that we do not need a data association process along video frames. This is because we associate the height priors with the image region sizes at image places where map features projections fall within the object detection regions. We present very promising results of this approach obtained on several experiments with different object classes.
	\end{abstract}

	\section{Introduction}
	
	Live monocular Simultaneous Localization and Mapping (SLAM) is a classical problem in computer vision, with many proposed solutions throughout the last decade~\cite{Davison2003,Klein2007}. It has numerous applications, from augmented reality to map building in robotics. However, when performing a 3D reconstruction with a calibrated, monocular system, it is well known that the output reconstruction is an estimate of the real structure up to an unknown scale factor. This may be a problem when the considered application involves measuring distances, inserting real-size virtual objects with a coherent scale, etc. Of course, in some specific contexts (in mobile robotics, for example), the unknown scale factor can be recovered easily by integrating information from other sensors~\cite{IMU, kinectfusion}, from some assumptions on the environment or on the position of the camera~\cite{Scaramuzza2009}, or from the introduction of known objects in the scene~\cite{GalvezLopez2016, Salas2013}.     
 \begin{figure}[t!]
	\begin{center}
 \includegraphics[width=\linewidth]{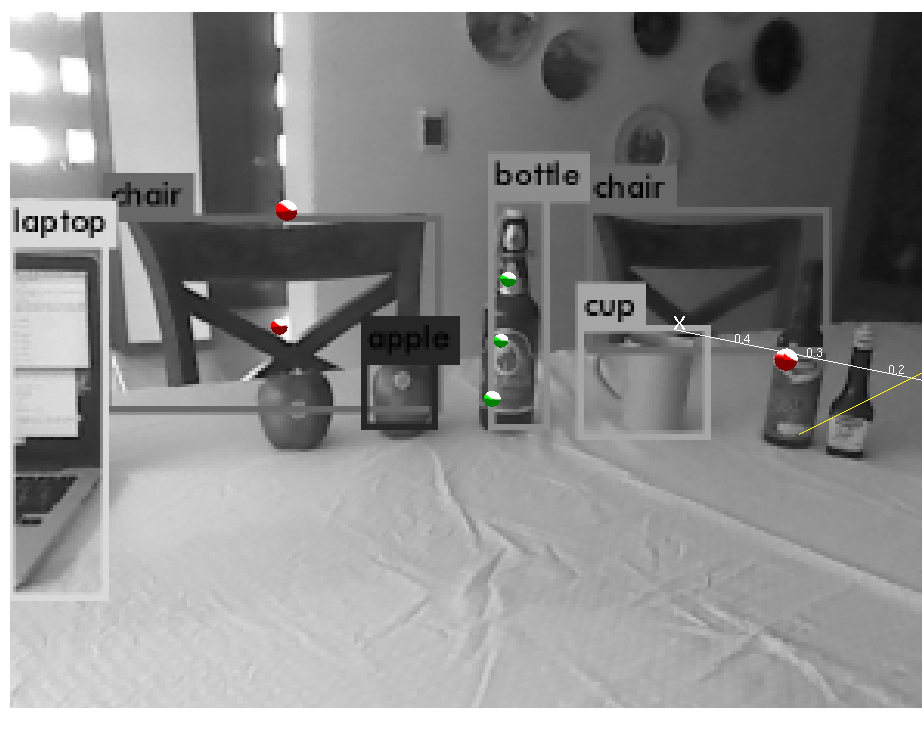}
	\end{center}
	\caption{Our approach combines the YOLO object detection algorithm running simultaneously with the MonoSLAM algorithm. The global map scale can be inferred from 3D map features whose projection (dots) are inside one of the object region  (rectangles) by incorporating the object height prior for that object class.\label{im_intro}}
\end{figure}
	
	Nevertheless, one can note that this intrinsic limit of monocular systems does not prevent animal visual systems to have a rather precise depth perception, even with one sensor only, and in very general situations. This perception of depth based only on monocular cues is rather well documented~\cite{perception}. In computer vision, the idea of inferring depth from monocular cues has been exploited in previous works relating monocular perception of texture and depth through machine learning techniques~\cite{Saxena2009}, and it has been used to initialize the unknown global scale in monocular SLAM systems~\cite{Mota2013}. In this work, the motivation is to follow this line of research from a different, much less explored perspective. We use region-based, semantic information instead of pixel-based information, and rely on the notion of familiar size monocular cues, well studied in psychology~\cite{perception}. To achieve this goal, as illustrated in Figure~\ref{im_intro}, we take advantage of the recent availability of very efficient techniques to detect instances of classes of objects (``bottle'', ``mug'', ``book'', etc.), made possible by the advent of deep learning techniques, and use an out-of-the-box generic object detector to extract observations from images of the video sequence. By associating size priors to these object classes, we are finally able to infer the global scale of a monocular SLAM system, within a Bayesian framework. 
	
	To the best of our knowledge, this is the first work that couples tightly the detected instances of an object class detector with a probabilistic monocular SLAM system to perform live global scale estimation based on semantic information given by the detector, on a single frame basis, and in a Bayesian scheme. We did not develop the detector by ourselves, but relied on recent proposals from the deep learning community. In essence, our approach is limited to detections from classes that the detector has learned.
	
	The structure of this document is the following. In Section~\ref{sec-related} we review related work on scale estimation for monocular SLAM. In Section~\ref{sec-method} we give an overview of our estimation method, and in Section~\ref{sec-likelihood} we present the likelihood model to integrate the object detection observations in the scale estimation framework. Section~\ref{sec-impl} gives an overview of the implementation, and in Section~\ref{sec-results} we present the experimental results and evaluations of these results with respect to simpler forms of estimating the unknown global scale. Finally, we summarize our work and discuss potential improvements in Section~\ref{sec-discussion}.
    
	\section{Related Work}
	\label{sec-related}
	Monocular algorithms are widely used for 3D scene reconstruction and camera tracking. The reason for this is that they are cost-effective (only an inexpensive camera is required) and current methods are robust. There are mainly two approaches for doing this: optimization-based methods such as PTAM~\cite{Klein2007} and filtering-based methods such as MonoSLAM~\cite{Davison2003}. We have considered here the filtering-based approach, MonoSLAM, as it is of our interest to incorporate in our method the uncertainty associated with 3D feature locations.
	
	In monocular 3D reconstruction, the scale of the map is inherently unknown. Methods such as MonoSLAM and PTAM use a special initialization procedure of the system to set up the scale of the map. MonoSLAM uses a known object for system initialization and for fixing the scale of the map, which introduces the inconvenience of needing a specific object each time it is executed. PTAM assumes an initial known translation, which does not give a very precise scale initialization. Both of these methods also have in common the problem of scale drift due to inaccuracies in 3D reconstruction and tracking. 
	
	Two main techniques have been used for automatic scale estimation during 3D reconstruction. The first technique consists of using sensors that intrinsically obtain measurements that allow scale to be estimated, such as depth of features using Kinect~\cite{kinectfusion} or translation measurements using IMU sensors~\cite{IMU}. However, as they incorporate additional sensors, these techniques defeat one of the main advantages of monocular algorithms which is the possibility of relying only on an inexpensive RGB camera. The second family of techniques consists of using known ``natural'' objects to estimate the scale of the map. The disadvantage of these techniques is the need to train an object recognition algorithm for specific objects and to have these objects introduced in the scenes in which the 3D reconstruction algorithm will be run. 
    
    In~\cite{Salas2013}, real-time 3D object recognition is used in a live dense reconstruction method based on depth cameras, to get 3D maps with a high level of compression and  complementary interactions between the recognitions, mapping and tracking processes. A major difference with our system is that it considers a database of fixed, specific objects, whereas our aim is to handle fixed categories of objects.

The use of semantics through objects in the SLAM context is also present in works similar to ours, as~\cite{GalvezLopez2016}, where an object recognition system, driven by a bag-of-words algorithm, is integrated within an object-aware SLAM system to impose constraints on the optimization. Again, a database of pre-defined specific objects is used, while our approach relies on much broader object categories.
    
    An alternative approach has been developed with the idea of using more generic objects, namely faces, using the front cellphone camera~\cite{ismarscale}. This method requires a special routine to be done during the 3D reconstruction by the back cellphone camera, in which 6DoF face tracking with the front facing camera with  known scale  is used  to obtain the scale  for the 3D reconstruction. This method does not generalize easily to other object classes as it depends on the precise 6DoF tracking algorithms for faces.
    
        In~\cite{Frost2016}, the most similar work to this one, a generic detector is also used, in the context of urban scenes, and priors on the sizes of the detected objects. However, a first difference to our work is that the reconstruction is done in a bundle adjustment framework and, above all, no connection is done between the reconstructed map and the detected object, so that the scale inference is done by using data association through consecutive frames, which is done prior to running the method. Our method does not require data association, as map features and detection regions are associated naturally on single frames. The Bayesian framework of our method also has the advantage that it is easy to incorporate additional uncertainties other than 3D feature locations, such as the uncertainty in the fit of the object's bounding box.
        
	Our approach aims to be ubiquitous for 3D reconstruction. It avoids the use of external sensors and it is robust, as it is based on a probabilistic framework. It uses a generic object recognition algorithm which runs in real time. Deep neural networks have achieved good accuracy for generic object recognition~\cite{RCNN}, and recently methods that run in real time have been developed~\cite{yolo}. By combining probabilistic height priors for recognized generic objects with 3D reconstruction uncertainties within a Bayesian framework, we are able to estimate the scale for monocular SLAM. 
	
	\section{Overview of our approach}
	\label{sec-overview}
	
	\label{sec-method}
	A monocular SLAM, such as MonoSLAM~\cite{Davison2003}, allows to reconstruct a scene as a sparse cloud of 3D points $\mathcal{N}=\{\mathbf p_i\}_{1\leq i \leq N}$, where
	
	\begin{equation}
	\mathbf p_i = (x_i,y_i,z_i)^T.
	\end{equation}
	
	It also allows to track the configuration of the camera along time $\mathbf x_{v,k}$, where $k$ is an index corresponding to time/camera frame and $\mathbf x_v$ is a representation of the 3D configuration of the camera, i.e., an element of the Special Euclidean group, $SE(3)$. The vision algorithms reconstruct the scene up to an unknown scale factor, this means that the \textbf{true} 3D points $\mathbf q_i$ are related to $\mathbf p_i$ through
	
	\begin{equation}
	\mathbf q_i = d \mathbf p_i + \nu,
	\end{equation}
	
	\noindent
	where $d$ is a global scale factor for the whole scene, {\em a priori} unknown, and  $\nu$ is the reconstruction error noise. Our aim is to estimate $d$ using Bayesian inference, based on object detections given by a generic object detector. 
	
	We take the approach proposed in~\cite{Civera07} to separate the state vector (3D coordinates of the map points and camera pose) into: (i) a dimensionless part that can be maintained based on the perspective projections equations as in~\cite{Davison2003} and, (ii) a recursive estimation scheme that maintains an estimate of the global scale $d$ as explained hereafter. 
	
	Then, our idea is to use those 3D points reconstructed by the SLAM algorithm that project into an ``object" region. In some way, the data association between the objects and the current reconstruction is done implicitly at these points. Let $D_k$ be the number of objects detected by the detector, and $\mathcal{D}^k=\{S_l\}_{1\leq l \leq D_k}$ be the set of $D_k$ detected object zones (bottles/chairs/screens...). Each $S_l$ encodes a detection performed in the image by the detector, and is associated to a rectangular window in the video frame $k$, containing a detected object, and a class $c_l$ for the detected object.
	
	We handle a subset of 3D points $\mathcal{T}_k\subset \mathcal{N}$, a subset of $T_k$  points reconstructed up to time $k$, characterized by the fact that their projection lies within an ``object" region as specified by the detector at time $k$. Let $\mathbf p_i \in \mathcal{T}_k$ be one of these points.  We will note at time $k$:
	
	\begin{itemize}
		\item $\mathbf{\pi}^k_i$ the 2D projection of $\mathbf p_i$ on image $k$.
		\item $H_l$ the function that associates one ``object" region to its real height, which we do not know in general, but for which we will have an associated prior.
        \item $c_l$ the function that associates one ``object" region to the index of the object class to which it belongs.
	\end{itemize}
	
	We are given a prior distribution on the possible heights for each object class $c$, i.e., a distribution, $p_{c}(H)$, that we build beforehand. For the moment, this distribution is set arbitrarily (see the Experimental Results section) but we plan to learn it by using large categories databases.
    
	Now, what we want to estimate is, at time $k$, the posterior probability for the coefficient $d$, i.e., the distribution conditioned on the cloud reconstructed up to time $k$ and the different detection-observations gathered: $p(d|\mathcal{N},\mathcal{D}^1,\dots,\mathcal{D}^k).$
	
	For the sake of clarity, we will suppose that $D_k=1$, so that there is only one object, with its detection $S_1$, and with its associated height prior $p(H_1)$. Then, by using Bayes rule, we can write the global scale posterior into
	\begin{small}
		\begin{equation}
		\begin{split}
		& p(d|\mathcal{N},\mathcal{D}^1,\dots,\mathcal{D}^k) \\ 
		& \propto 
		p(\mathcal{D}^k|d,\mathcal{N},\mathcal{D}^1,\dots,\mathcal{D}^{k-1})p(d|\mathcal{N},\mathcal{D}^1,\mathcal{D}^2,\dots,\mathcal{D}^{k-1})\\
		& \propto 
		p(S_1|d,\mathcal{N},\mathcal{D}^1,\dots,\mathcal{D}^{k-1})p(d|\mathcal{N},\mathcal{D}^1,\mathcal{D}^2,\dots,\mathcal{D}^{k-1})\\
		& \propto    p(d|\mathcal{N},\mathcal{D}^1,\dots,\mathcal{D}^{k-1}) \\ & \int_{H_1} p(H_1,S_1|d,\mathcal{N},\mathcal{D}^1,\dots,\mathcal{D}^{k-1}) dH_1 \\
		& \propto    p(d|\mathcal{N},\mathcal{D}^1,\dots,\mathcal{D}^{k-1}) \int_{H_1} p(S_1|H_1,d,\mathcal{N},\mathcal{D}^1,\dots,\mathcal{D}^{k-1}) \\ & p(H_1|d,\mathcal{N},\mathcal{D}^1,\dots,\mathcal{D}^{k-1})dH_1  \\
		& \propto    p(d|\mathcal{N},\mathcal{D}^1,\dots,\mathcal{D}^{k-1}) \int_{H_1} p(S_1|H_1,d,\mathcal{N}) p_{c_1}(H_1)dH_1.  
		\end{split}
		\end{equation}
	\end{small}
	
	This means that at each step we can update the posterior on $d$ by multiplying it by $\int_{H_1} p(S_1|H_1,d,\mathcal{N}) p_{c_1}(H_1)dH_1$. We approximate this term  with the help of a histogram representation over $H_{1}$, $\{p_{c_1}(H_{1,m})\}$, and the prior probability that a detection of object $1$, of some specific class, has a real height $H_{1,m}$ (where the heights have been discretized): $\sum_{m} p(S_1|H_{1,m},d,\mathcal{N}) p_{c_1}(H_{1,m}).$
	
	In the more general case of $D_k>1$, and by assuming conditional independence between the different detections observed in frame $k$, one can show that a similar development leads to  
	
    \begin{equation}
\begin{split}	&p(d|\mathcal{N},\mathcal{D}^1,\dots,\mathcal{D}^k)\\ 
&= p(d|\mathcal{N},\mathcal{D}^1,\dots,\mathcal{D}^{k-1}) \prod_{l=1}^{D_k} \int_{H_l}  p(S_l|H_l,d,\mathcal{N}) p_{c_l}(H_l)dH_l.
\end{split}
	\end{equation}
	
	The likelihood term $p(S_l|H_{l,m},d,\mathcal{N})$ is the probability that the detected object has the dimensions in pixels with which it was detected, given that the object has some real size $H_{l,m}$, that the scale is $d$, and that the cloud is $\mathcal{N}$. Its calculation is explained in the next section.  
	
	The evaluations over $d$ are done for a discrete set of values, $d_i$, within some specified interval $[d_{min},d_{max}]$, to ease the computational burden.
	
	\section{Likelihood of the observations}
	\label{sec-likelihood}
   We will now describe the calculation of the likelihood term $p(S_l|H_{l,m},d,\mathcal{N})$, which is the probability that the detected object has the dimensions in pixels with which it was detected, given that the object has some real size $H_{l,m}$, that the scale is $d$, and that the 3D cloud is $\mathcal{N}$. The  general idea for doing this is back projecting the object extremities found in the image into the 3D map and estimating the object's height using a feature in $\mathcal{N}$ that projects into the object and the scale $d$. The height estimate is then compared with $H_{l,m}$ to obtain  $p(S_l|H_{l,m},d,\mathcal{N})$.
   
	Let us assume that in the world  frame, i.e., the frame in which the 3D points are reconstructed by MonoSLAM, we know the vertical direction, that is, the direction perpendicular to the ground plane. This vertical direction is set when initializing the MonoSLAM system with a square marker which is perpendicular to the ground plane. Let also $\mathbf R^W_k$ be the 3D location of the camera at time $k$ in the world frame $W$. The vectors and points we will be dealing with hereafter will be expressed in the camera frame.
	
	\begin{figure}[t]
		\begin{center}
			\includegraphics[width=0.8\linewidth]{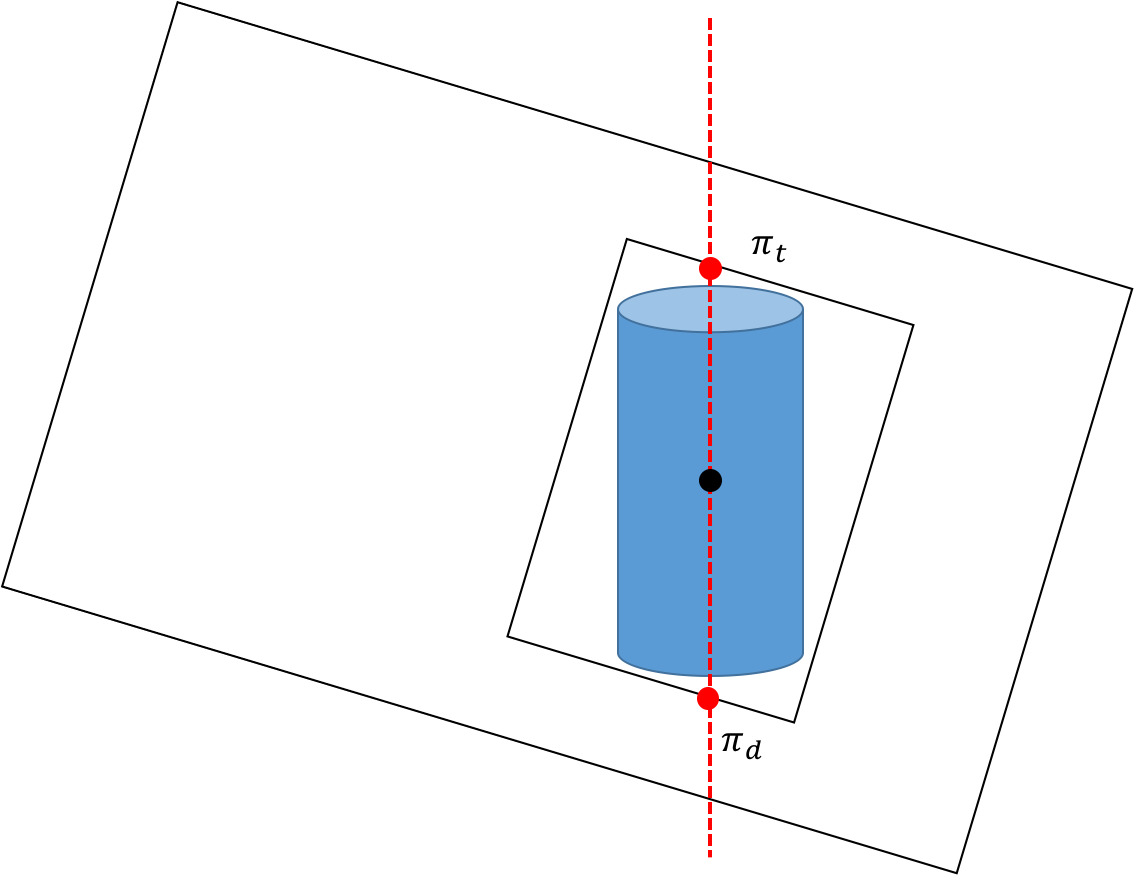}
		\end{center}
		\caption{Points of intersection, $\pi_t$ and $\pi_d$, of $\lambda$ with the boundary of the object detection.}
		\label{intersect}
	\end{figure}
	
\begin{figure}[t]
		\begin{center}
			\includegraphics[width=0.8\linewidth]{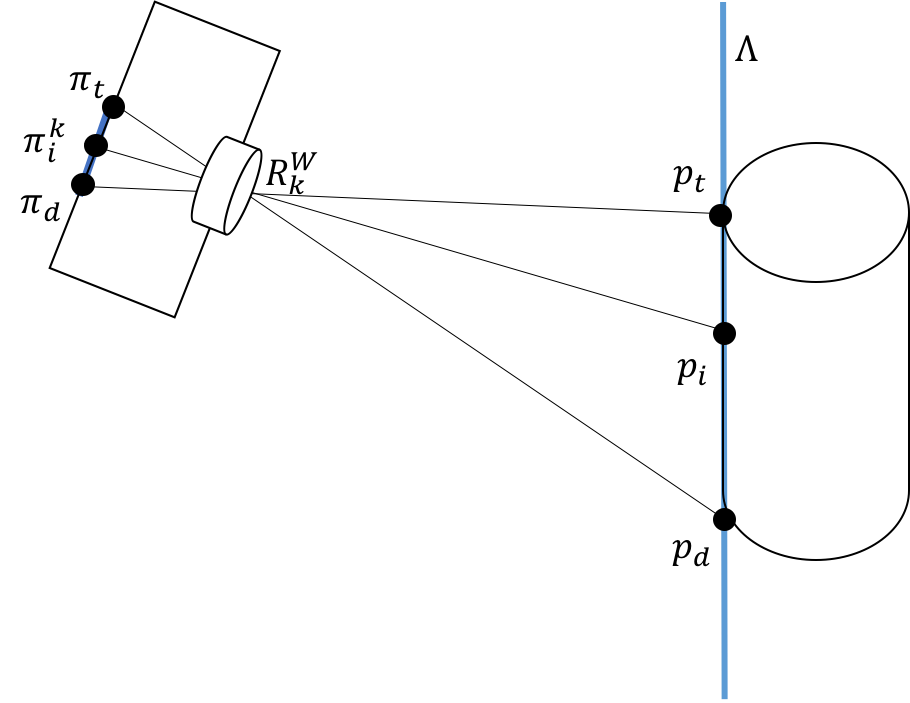}
		\end{center}
		\caption{Camera-Object configuration for point-cloud observations. $R_k^W$ is the 3D camera center, $\pi_t$ and $\pi_d$ are the object extremities in the image. $p_i$ is a 3D point in the object with $\pi_i^k$ its image projection. $p_t$ and $p_d$ are the 3D object extremities calculated using $\pi_t$, $\pi_d$, and the world frame vertical line $\Lambda$.}
		\label{ray_L}
	\end{figure}

	To evaluate $p(S_l|H_l,d,\mathcal{N})$, let us consider a 3D point $\mathbf p_i$ from the reconstructed cloud $\mathcal{N}$, such that its current projection $\mathbf{\pi}^k_i \in S_l^k$.  Let $\lambda$ be the line in  the image such that it contains  $\mathbf{\pi}^k_i$ and such that the plane obtained by back projecting this line onto the 3D map is parallel to the vertical direction.
	Let us define the intersections of this line with the boundary of the detection $S_l$, $\mathbf \pi_t, \mathbf \pi_d = \lambda \cap \partial S_l^k$, as depicted in Figure~\ref{intersect} with red dots, while $\mathbf{\pi}^k_i$ is the black dot. These two points in the image correspond to the vertical extremities of the object.  Now let us consider the line $\Lambda$ in the 3D map such that it contains $\mathbf p_i$ and that it is parallel to the vertical direction, i.e., we will suppose that the detected object of interest is aligned with the vertical direction in the world frame. Let $ \hat{\pi}_t$ and $ \hat{\pi}_d$ be the 3D map rays obtained by back projecting the image points $\mathbf \pi_t$ and $\mathbf \pi_d$, respectively. Now we define  $\mathbf p_t = \hat{\pi}_t \cap \Lambda$ and  $\mathbf p_d = \hat{\pi}_d \cap \Lambda$. These two points correspond to the vertical extremities of the object in the 3D map, as seen in Figure~\ref{ray_L}. Note that the coordinates of $\mathbf r^W_k$, together with the coordinates of the point $\mathbf p_i$, are given in the dimensionless state vector.  
	Then the estimated object height can be approximated as the Euclidean distance $D(\mathbf p_t,\mathbf p_d)$.
	
	Given $f$, a symmetric density centered at zero (e.g., a Gaussian), we can evaluate $p(S_l|H_l,d,\mathcal{N})$ as 
	
	\begin{equation}
	p(S_l|H_l,d,\mathcal{N}) =  f(|d D(\mathbf p_t,\mathbf p_d)-H_l|;0,\sigma).
	\end{equation}
	
	The dispersion parameter $\sigma$ in $f$ is important to define to give a proper weight to each observation in the estimation scheme. The density $f$ is modeled as a Gaussian density since the only source of uncertainty considered at the moment is the 3D position of $\mathbf p_i$, which is estimated through a Kalman filter. In particular, as the position of $\mathbf p_i$ is uncertain, and is estimated as in~\cite{Davison2003} in the dimensionless space, with an uncertainty $P_{p_ip_i }$, we can estimate the expected variance on $D(\mathbf p_t,\mathbf p_d)$. Let us associate the same covariance matrix $P_{p_ip_i }$ (calculated by the MonoSLAM algorithm) to each of the points $\mathbf p_t,\mathbf p_d$. Then, the variance on $D$ can be calculated by standard uncertainty propagation:
	
	\begin{equation}
	\begin{split}
	\sigma^2_D & =  \frac{\partial D}{\partial \mathbf p_t} P_{p_ip_i } (\frac{\partial D}{\partial \mathbf p_t})^T + \frac{\partial D}{\partial \mathbf p_d} P_{p_ip_i } (\frac{\partial D}{\partial \mathbf p_d})^T,\\
	&= 2 \frac{\partial D}{\partial \mathbf p_t} P_{p_ip_i } (\frac{\partial D}{\partial \mathbf p_t})^T\\
	&=2 J P_{p_ip_i } J^T,
	\end{split}
	\end{equation}
	
	\noindent
	where $J$ is the gradient of the distance function with respect to $\mathbf p_t$, i.e.,
	
	\begin{equation}
	J=\frac{1}{D}(\mathbf p_t-\mathbf p_d).
	\end{equation}
	
	Then, given that $f(d; 0, \sigma)$ is the density of a  Gaussian random variable with variance $\sigma^2$, $F(d) = f(|d D(\mathbf p_t,\mathbf p_d)-H_l|; 0, \sigma)$ is the density of a  Gaussian random variable with variance $\frac{\sigma^2}{ D(\mathbf p_t,\mathbf p_d)^2}$ and for that reason we use $\sigma^2=\sigma^2_D D(\mathbf p_t,\mathbf p_d)^2$ as the variance of $f$, so that $f(|d D(\mathbf p_t,\mathbf p_d)-H_l|; 0, \sigma)$ has variance $\sigma^2_D$.
	
The global state posterior is updated every time a new observation is obtained by multiplying it by the likelihood of the new observation: $\sum_{H_{l,m}} p(S_l|H_{l,m},d,\mathcal{N}) p(H_{l,m})$. An observation is generated by each 3D feature whose projection lies inside an object detection region each time the object detection algorithm is ran.
    The dispersion of the likelihood of the new observation is calculated using the covariance of its respective 3D feature as described above. We expect the likelihood to have a larger dispersion the bigger the covariance of the 3D feature is. The scale parameter $d$ is calculated as the mode of $p(d|\mathcal{N},\mathcal{D}^1,\dots,\mathcal{D}^k)$, the global state posterior (MAP). 
    
\begin{figure}[t]
		\begin{center}
			(a) \includegraphics[width=\linewidth]{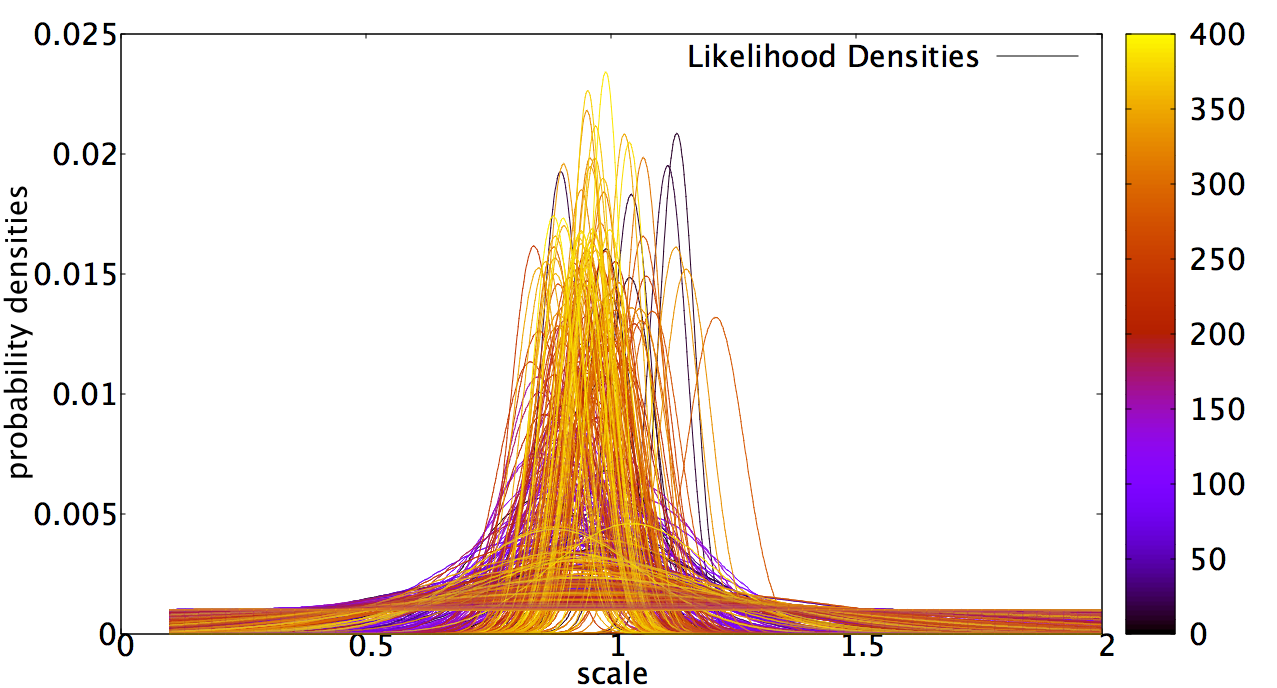}\\
			(b) \includegraphics[width=\linewidth]{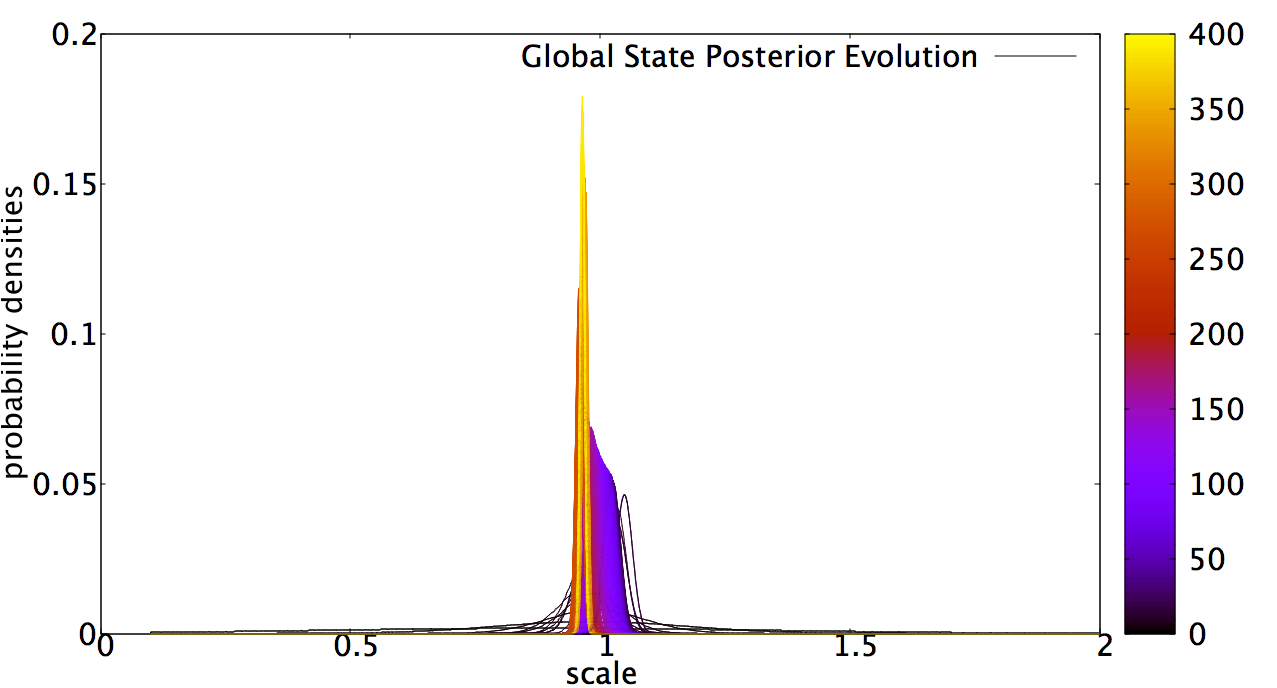}\\
			(c) \includegraphics[width=\linewidth]{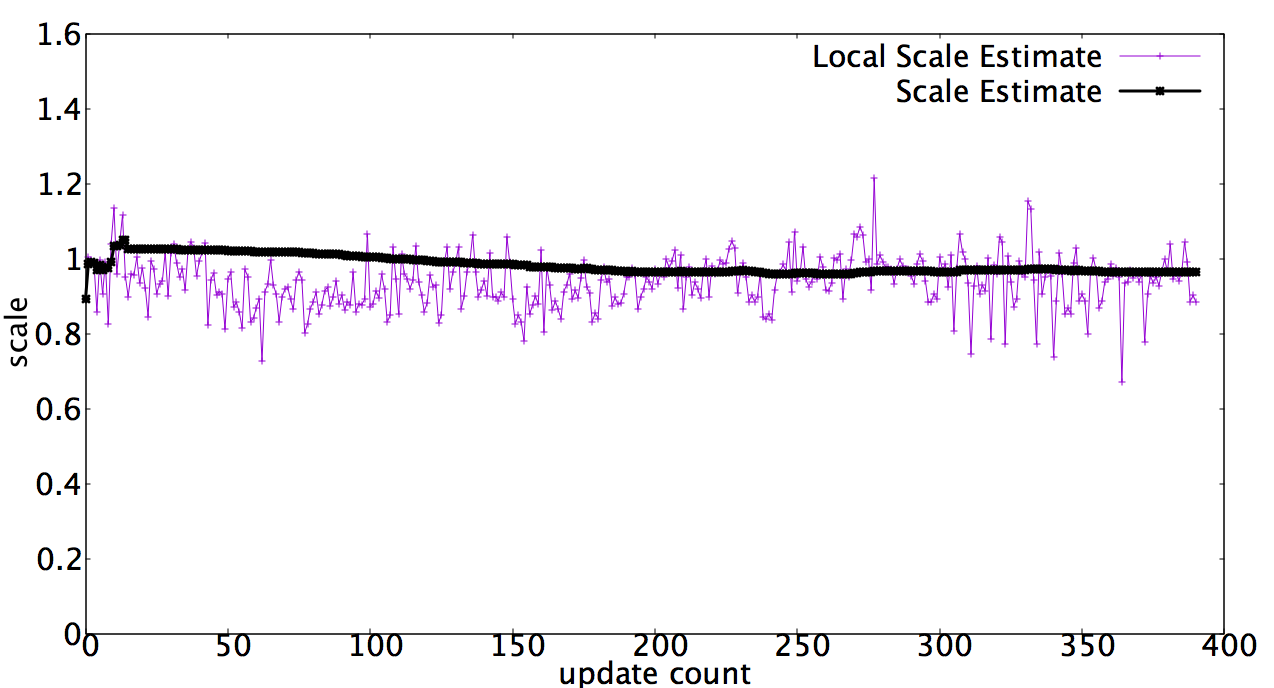}
		\end{center}
		\caption{(a) Likelihoods of  new observations. We observe likelihoods with different dispersions since the dispersion is calculated from the covariance of its respective 3D feature. The color palette indicates the point in time (the clearer the later in the video). (b) Evolution of the global scale posterior, and (c) evolution of the global scale MAP estimate (in bold) along local state estimates for each update (graphs correspond to experiment 1).}
		\label{density}
	\end{figure}

	We can also calculate a local scale estimate specific to each global state posterior update, as the mode of $L(d) = \sum_{H_{l,m}} p(S_l|H_{l,m},d,\mathcal{N}) p(H_{l,m})$, the likelihood of the new observation; this local scale estimate can be interpreted as an observation of the true scale. For instance, Figure~\ref{density} (where the graphs correspond to experiment 1 described in Section \ref{sec-results}) depicts in (b) the evolution along the video frames of the global scale posterior, with a clear tendency to reduce the scale estimate variance, in (a) the likelihoods of the new observations corresponding to each global state posterior update, and in (c) the evolution of the global scale parameter estimate along the the local estimates obtained after each update. The local scale parameter estimates allow us to observe the variability of the ``observations'' being made.
	
    		\begin{figure}[t]
		\begin{center} \includegraphics[width=\linewidth]{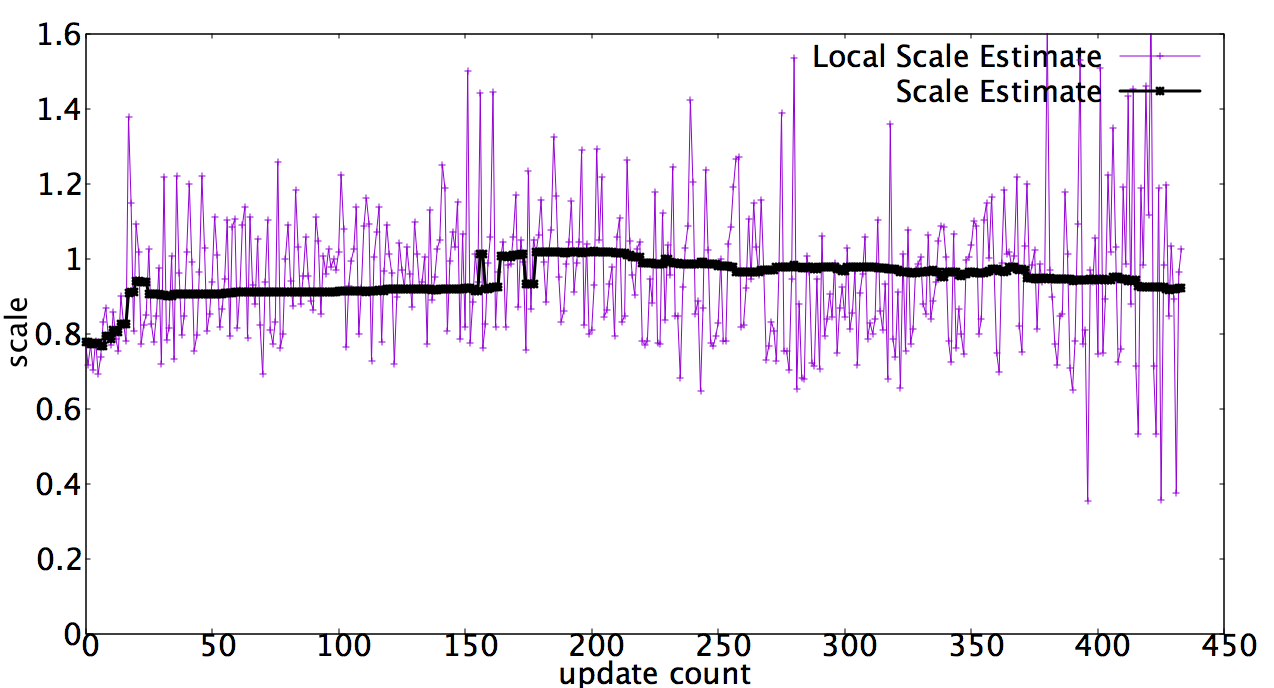}\\
		\end{center}
		\caption{Evolution of the global scale MAP estimate (in bold) along local state estimates for each update in experiment 3.}
		\label{scale3}
	\end{figure}

   \section{Implementation}
   \label{sec-impl}
   
	Our implementation uses the SceneLib2 library~\cite{SceneLib} which is a reimplementation of Davison et al.'s~\cite{Davison2003} original algorithm for Kalman-based Monocular SLAM. It also uses the YOLO v2 algorithm~\cite{yolo}, which runs on real time on an NVIDIA GPU, for generic object detection. YOLO v2 was trained on the 2007 and 2012 PASCAL Visual Object Classes Challenge datasets~\cite{pascal}. Both of these algorithms run on real time, and our method requires little computational expense, which guarantees a real time execution of the final algorithm. 
	
	The object detection function is run every 10 frames. Each feature point inserted in the map and whose projection falls inside an object region is used to update the global scale posterior as described in Section~\ref{sec-overview}.

\begin{figure}[t]
		\begin{center}
			(a) \includegraphics[width=0.8\linewidth]{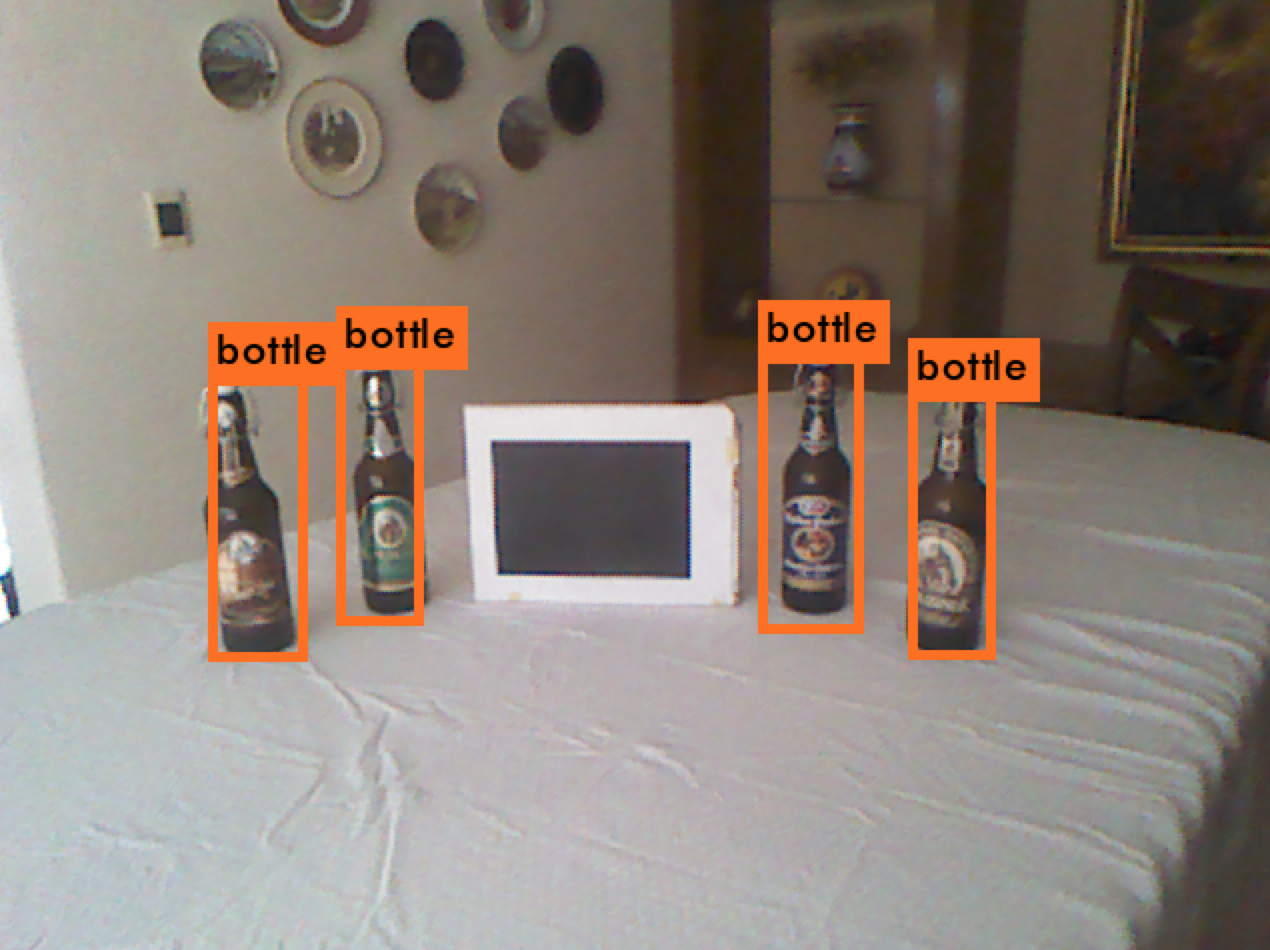}\\
			(b) \includegraphics[width=0.8\linewidth]{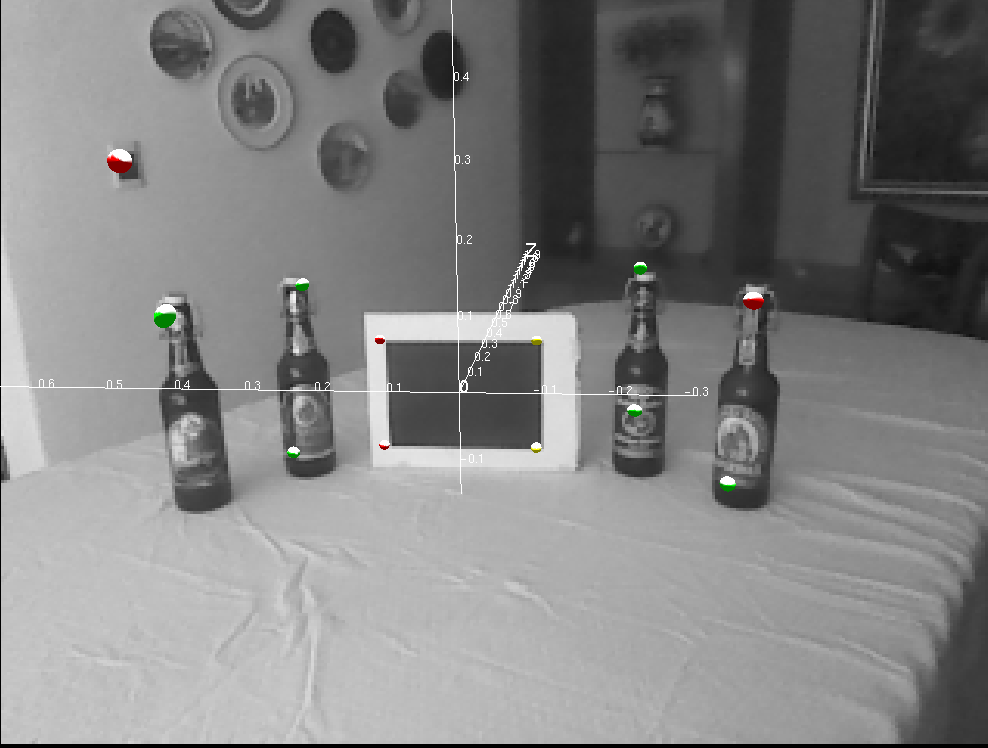}\\
			(c) \includegraphics[width=0.8\linewidth]{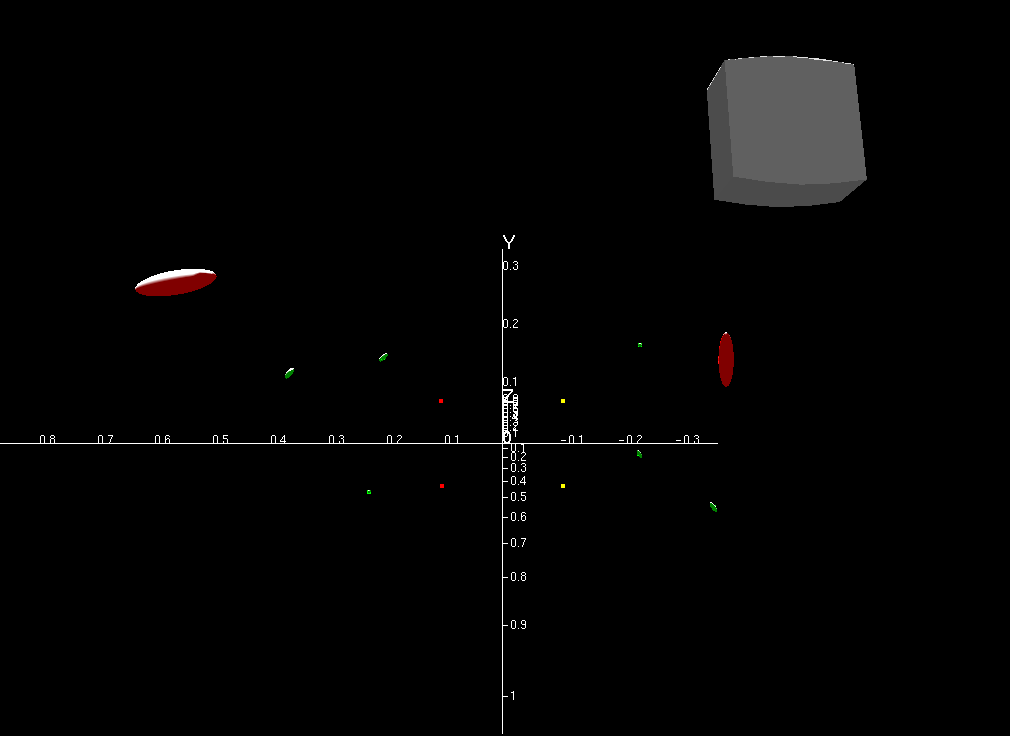}
		\end{center}
		\caption{(a) An instance of the YOLO v2 object (bottle) detection. (b) Feature points detected, those shown in green are inside an object detection region (the detected bottles).  (c) 3D Map depicting the feature points and camera position (images correspond to experiment 1). }
		\label{impl}
	\end{figure}

    Figure~\ref{impl} shows an instance of the algorithm for an specific frame, showing  the object detection result, (a), and camera tracking with 3D map features, (b) and (c). 
	
	\section{Experimental Results}
	\label{sec-results}
	
We first describe the evaluation methodology and then present the quantitative results.

	\subsection{Evaluation method}

The evaluation of the method is run in parallel to the algorithm by comparing distances computed based on the scale estimate with distances obtained by a Kinect. For this comparison, we use a marker (the same that is used to initialize the MonoSLAM system) with four unambiguous 3D map features. Let $y_1^W$, $y_2^W$, $y_3^W$, and $y_4^W$ be the locations of these features as estimated by the MonoSLAM algorithm. Each time the global state posterior is updated, a new scale parameter is estimated and this error is computed for this scale parameter.  Let $D_R()$ be a function that measures the true distance from the camera to each feature obtained with a Kinect. For each feature, an absolute error is computed as:
\begin{equation}
e_i(d) = |D_R(y_i^W)  - d D(y_i^W, R_k^W)|,
\end{equation}
with $R^W_k$ the 3D position of the camera at the current frame $k$.
Given that there are four feature points, the total absolute error is calculated as:
\begin{equation}
\epsilon(d)=\frac{1}{4}\sum_{i=1}^4 e_i(d).
\end{equation}

The relative error  for each feature is defined as:
\begin{equation} 
\delta_i(d) = 100 \frac{|D_R(y_i^W)  - d D(y_i^W, R_k^W)|}{D_R(y_i^W)}.
\label{rel_error}
\end{equation}
The total relative error is computed as:
\begin{equation}
\Delta(d)=\frac{1}{4}\sum_{i=1}^4 \delta_i(d).
\end{equation}

\subsection{Results}

To evaluate the proposed method we performed three experiments considering different types and number of objects in the scene. In each case a we ran MonoSLAM and the object detector over a video sequence, estimating the scale with our method. The absolute and relative errors are estimated according the procedure described above. We report the median and standard deviation of these errors for all the scale updates after the scale has converged, which is determined visually, and show the evolution of the relative error. To have an indication on the expected fluctuations, the relative errors are also calculated without using the scale estimate, that is, setting $d=1$ in equation ~\ref{rel_error}, which maintains the initial scale that is set by the initialization of MonoSLAM with a marker of known dimensions. The purpose of this is determining the cause of fluctuations in the relative error of the scale.

\subsubsection{Experiment 1}
In this first experiment, 4 bottles of the same size are used. Our height prior assigns the true bottle height a probability of 1, zero elsewhere (i.e., the object size is known). The sequence that we used contains 950 frames, with a total of of 226 global scale posterior updates; the statistics are reported after update number 80. The median absolute error is $0.0191m$ with standard deviation of $0.0097m$; the median relative error is $1.7197 \%$ with a standard deviation of $0.8621 \%$. Figure~\ref{error1} (a) shows the evolution of the relative error of the scale, and (b) the evolution of the relative error with our estimated scale vs. the evolution of the relative error with a scale set to 1.

\subsubsection{Experiment 2}
\begin{figure}[t]
	\begin{center}
 (a)\includegraphics[width=\linewidth]{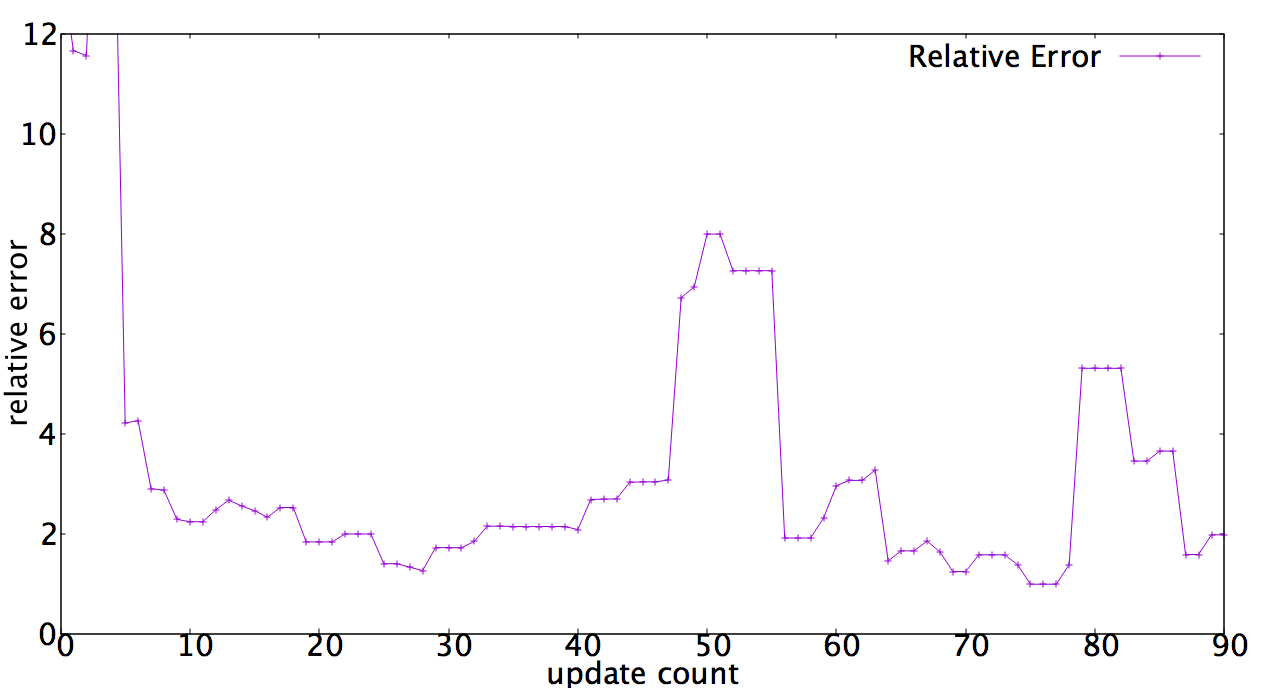}
 (b)\includegraphics[width=\linewidth]{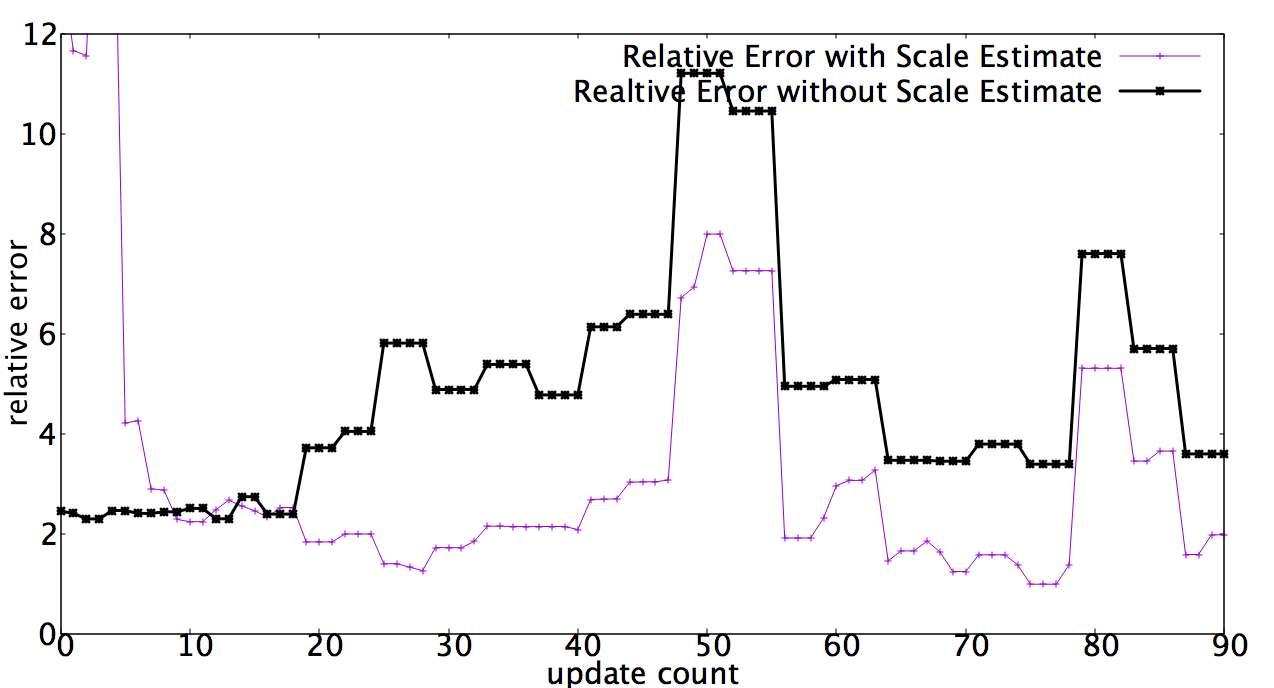}
	\end{center}
	\caption{(a) Evolution in time of the relative error of the scale for Experiment 2. (b) Evolution in time of the relative error of the scale, along the relative error for a constant scale  (in bold) for Experiment 2.}
	\label{error2}
\end{figure}

In our second experiment, another object (a microwave oven) is used. The height prior assigns the true microwave height a probability of 1 (i.e., the oven size is known). The sequence that we used contains 1073 frames, with a total of of 92 global state posterior updates; the statistics are reported after update number 16. The median absolute error is $0.0243m$ with standard deviation of $0.0171m$; the median relative error is $2.1455 \%$ with standard deviation of $1.7411\%$.
Figure~\ref{error2} (a) shows the evolution of the relative error of the scale, and (b) the evolution of the relative error with our estimated scale vs. the evolution of the relative error with a scale set to 1.

\subsubsection{Experiment 3}
\begin{figure}[t]
	\begin{center}
 (a)\includegraphics[width=\linewidth]{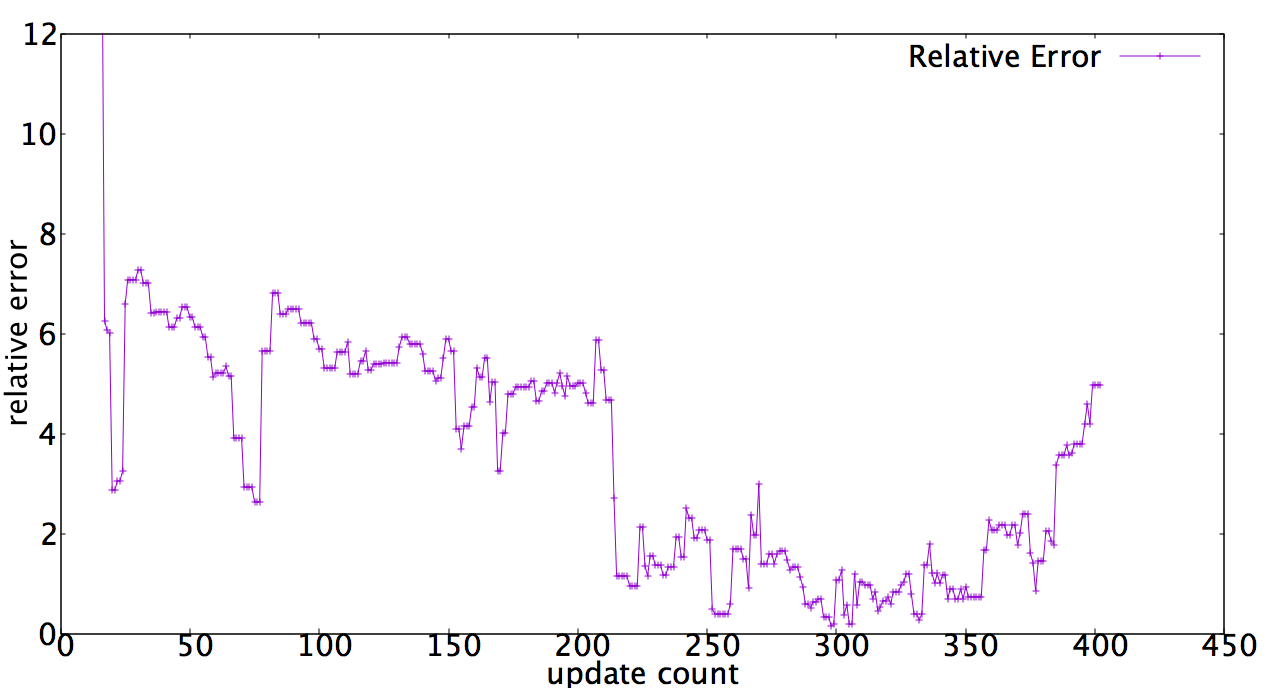}
  (b)\includegraphics[width=\linewidth]{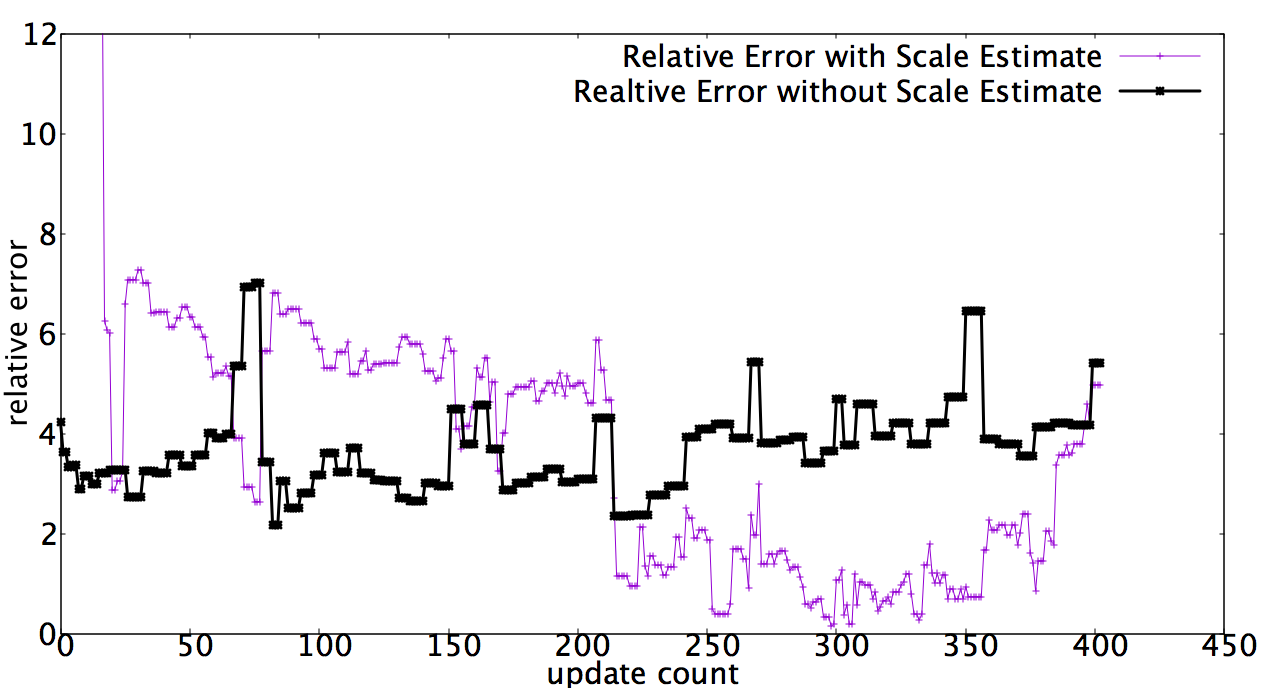}
	\end{center}
	\caption{(a) Evolution in time of the relative error of the scale for Experiment 3. (b) Evolution in time of the relative error of the scale, along the relative error for a constant scale  (in bold) for Experiment 3.}
	\label{error3}
\end{figure}

In our third experiment, 4 bottles of different sizes are used. For each object (bottles), the height prior is defined as a uniform histogram assigning to each of the four different bottle sizes a probability of $\frac{1}{4}$. The sequence that we used contains 1081 frames, with a total of of 403 global scale posterior updates; the statistics are reported after update number 215. The median absolute error is $0.0123m$ with a standard deviation of $0.0095m$; the median relative error is $1.3390 \%$ with standard deviation of $1.0266\%$. 
Figure~\ref{error3} (a) shows the evolution of the relative error of the scale, and (b) the evolution of the relative error with our estimated scale vs. the evolution of the relative error with a scale set to 1.

		\begin{figure}[t]
	\begin{center}
 (a)\includegraphics[width=\linewidth]{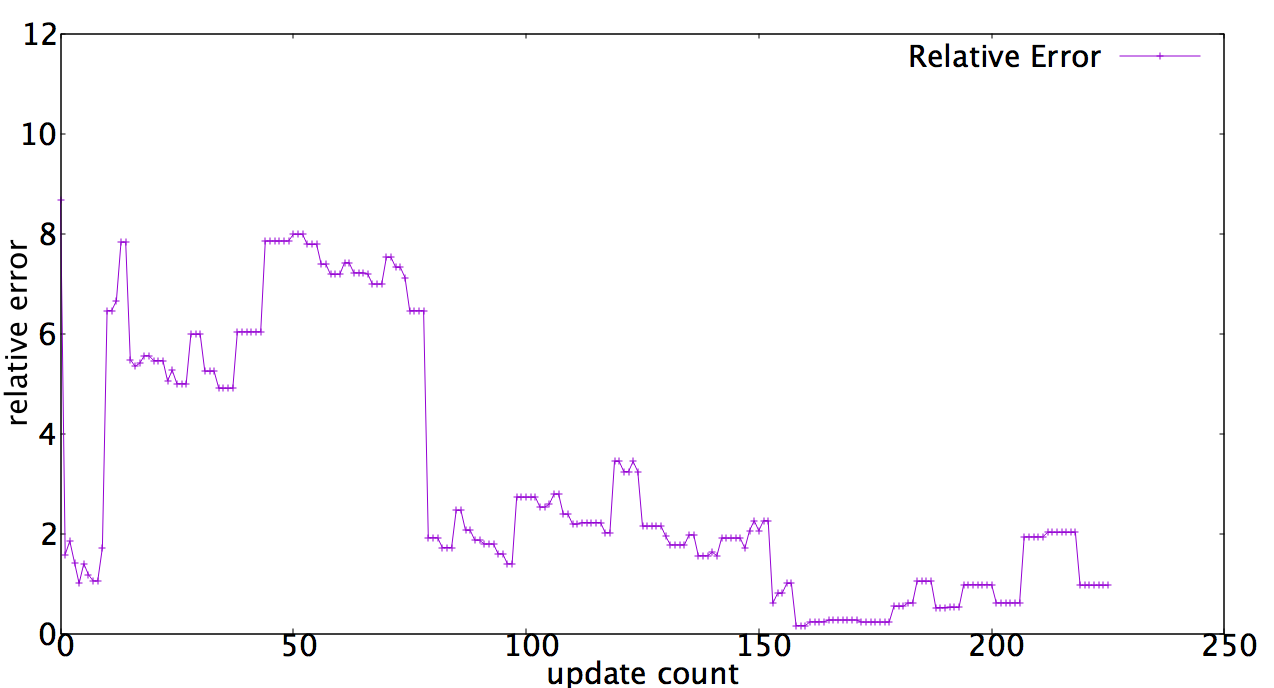}
 (b)\includegraphics[width=\linewidth]{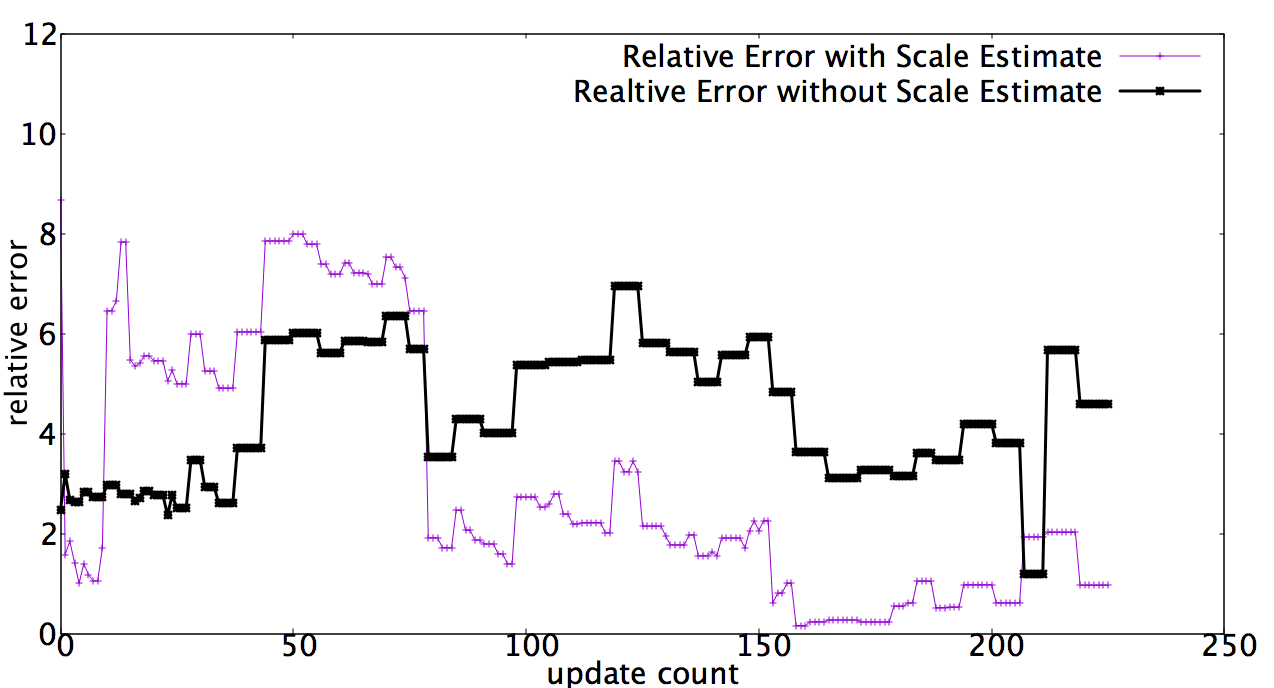}
	\end{center}
	\caption{(a) Evolution in time of the relative error of the scale for Experiment 1. (b) Evolution in time of the relative error of the scale, along the relative error for a constant scale  (in bold) for Experiment 1.}
	\label{error1}
\end{figure}

	\subsection{Discussion}

We observe in the three experiments how the error decreases as more object observations are integrated in the scale estimate, obtaining in all cases a median relative error very close to 2\%, after the scale estimate has converged. As expected, when there is a larger variance in the heights of the objects (Experiment 3), the error takes more updates to converge. However, even in this case, the median error is very low after convergence, with a median of  $1.3390 \%$.
    
    In the three experiments we can observe a bigger error in the first few updates of the scale. The reason for this is the large covariance of the 3D features used to estimate the scale, since the MonoSLAM algorithm is at its initial stage.
    
    It may appear that the scale estimate is not stable when observing the graphs of the evolution of the relative error. However, when analyzing Figure~\ref{density} and ~\ref{scale3}, the scale appears rather stable. These fluctuations present in the relative error appear to be due uncertainties from the Kinect measurements and the MonoSLAM position estimates. This is verified by comparing the evolution of the relative error with our estimated scale vs. the evolution of the relative error with a fixed scale of $1$ (keeping the initial scale from MonoSLAM, which is fixed by the features inserted on a marker with known dimensions), as can be seen in Figures  ~\ref{error1} (b), ~\ref{error2} (b), and ~\ref{error3} (b), where the fluctuations appear greatly correlated. 
    
\section{Conclusions and Future Work}
\label{sec-discussion}
    
We have developed a novel method to estimate the global scale of a 3D reconstruction for monocular SLAM. Based on the recent advances in generic object recognition, we use a Bayesian framework to integrate height priors over observations of detected object sizes in single frames. The method does not require temporal data association as map features and detection regions are associated naturally on single frames. The proposed approach takes advantage of recently developed techniques for object recognition based on deep learning that run in real time.
    
Experimental results considering different number and types of objects give evidence of the feasibility of our approach, obtaining median relative errors in average of 2 \%. These preliminary experiments show that it is possible to perform a rather precise global scale estimation based on priors on semantic classes of objects.  This precision could allow applications in Augmented Reality to insert objects in a scene with coherent size or scale drift correction for navigation of autonomous cars and drones.
    
    There are a few limitations to our approach that we are currently searching to improve. In particular, the map features projections may fall on the detection regions while the corresponding 3D ray does not intersect the 3D object. An outlier rejection scheme could be used to filter out these observations.  Another option would be to use a semantic segmentation algorithm for a more precise outline of the object to avoid  3D map features not in the object falling in the detection outline.
    
    We have used object classes with relatively small intra-class size variance; we need to test this approach on more complex priors.  We intend to develop a methodology for constructing prior height distributions for different classes of objects using measuring sensors such as Kinect, sampling a wide range of specific objects for each given class.
    
	As the proposed method is mostly independent of the 3D reconstruction and tracking algorithm, it should allow us to migrate the approach to  state of the art monocular SLAM systems as a future work.  This would allow us to test the method in public datasets such as the KITTI dataset \cite{Geiger2013IJRR} for a precise comparison with other methods for scale estimation.


	{\small
		\bibliographystyle{ieee}
		\bibliography{egbib}
	}
	
\end{document}